\documentclass{article}
\usepackage{graphicx} 
\usepackage[final]{neurips2024}

\usepackage{amsmath}
\usepackage{tcolorbox}
\usepackage{microtype}
\tcbuselibrary{skins}
\tcbuselibrary{raster}
\tcbuselibrary{breakable}
\usepackage{subcaption}
\usepackage{comment}

\usepackage{xcolor}

\usepackage{algorithm}
\usepackage{algorithmicx}
\usepackage{algpseudocode}
\usepackage{colortbl}
\definecolor{crimson}{HTML}{8B0000}

\usepackage{hyperref}
  \hypersetup{
    colorlinks=true,            
    linkcolor=crimson,          
    citecolor=crimson,          
    urlcolor=crimson,           
  }
\graphicspath{{figures/}}
\usepackage{natbib}
\title{Predictive Scheduling for Efficient Inference-Time Reasoning in Large Language Models}

\author{
  \textbf{Katrina Brown}\thanks{Equal contribution.} \\
  Harvard College \\
  \href{mailto:katrinabrown@college.harvard.edu}{katrinabrown@college.harvard.edu}
  \and
  \textbf{Aneesh Muppidi}\footnotemark[1] \\
  Harvard College \\
  \href{mailto:aneeshmuppidi@college.harvard.edu}{aneeshmuppidi@college.harvard.edu}
  \and
  \textbf{Rana Shahout} \\
  Harvard SEAS \\
  \href{mailto:rana@seas.harvard.edu}{rana@seas.harvard.edu}
}

\date{April 2025}

\begin{document}

\maketitle
\begin{abstract}
    Large language models (LLMs) achieve state-of-the-art accuracy on complex reasoning tasks by generating multiple chain-of-thought (CoT) traces, but using a fixed token budget per query leads to over-computation on easy inputs and under-computation on hard ones. We introduce Predictive Scheduling, a plug-and-play framework that pre-runs lightweight predictors—an MLP on intermediate transformer hidden states or a LoRA-fine-tuned classifier on raw question text—to estimate each query’s optimal reasoning length or difficulty before any full generation. Our greedy batch allocator dynamically distributes a fixed total token budget across queries to maximize expected accuracy. On the GSM8K arithmetic benchmark, predictive scheduling yields up to 7.9 percentage points of absolute accuracy gain over uniform budgeting at identical token cost, closing over 50\% of the gap to an oracle with perfect foresight. A systematic layer-wise study reveals that middle layers (12–17) of the transformer carry the richest signals for size estimation. These results demonstrate that pre-run budget prediction enables fine-grained control of the compute–accuracy trade-off, offering a concrete path toward latency-sensitive, cost-efficient LLM deployments.
    \\
    \\
    \vspace{0.5em}
    \noindent\fcolorbox{crimson}{crimson!10}{\parbox{0.95\linewidth}{\centering\small \textbf{Project Website \& Code:} \href{https://aneeshers.github.io/predictive-scheduling/}{https://aneeshers.github.io/predictive-scheduling/}}}
\end{abstract}

\section{Introduction}

Modern large language models (LLMs) deliver exceptional chain-of-thought (CoT) reasoning capabilities, powering applications from real-time code autocomplete to interactive tutoring and decision support. Yet inference-time costs remain a critical bottleneck: applying a fixed token budget per query either wastes tokens on trivial inputs or starves hard ones, leading to unnecessary latency and inflated cloud bills—key concerns for production LLM services and on-device deployments.

We first ask the problem of \emph{per-query budget heterogeneity} by asking: \emph{can we predict, before any generation, how many tokens each reasoning trace needs, or the difficulty of each query, in order to guide budget allocation?} Prior approaches either rely on few-shot heuristics to guess a single optimal trace length \citep{han2024token}, perform single-checkpoint early-termination checks during generation \citep{li2024escapeskyhighcostearlystopping, fu2024efficient}, or schedule batches based on surface-level signals such as queue length or past runtimes \citep{liu2023vllm, damani2024learninghardthinkinputadaptive}. None leverage the rich internal features of transformer hidden states to learn fine-grained, pre-run estimates of required trace length and query difficulty, nor integrate these estimates into a global batch allocator under a fixed total token budget.

Concretely, we present five contributions
\begin{enumerate}
  \item We introduce 2 lightweight predictors that map intermediate transformer hidden states to an estimate of the \emph{token length} required for a reasoning trace to achieve correctness.
  \item We develop a lightweight \emph{difficulty classifier}—using both few-shot prompting and LoRA fine-tuning—to categorize queries as easy, medium, or hard before generation.
  \item We perform a systematic layer-wise analysis, revealing that middle transformer layers (12–17) carry the strongest predictive signal for reasoning-length estimation.
  \item We design and implement a greedy batch allocation algorithm that dynamically assigns per-trace token budgets to maximize expected accuracy gains under a fixed total budget.
  \item We demonstrate on the GSM8K arithmetic reasoning benchmark that our combined predictive scheduling yields up to 7.9\% absolute accuracy improvement at equal token cost relative to nonadaptive scheduling, closing over half the gap to an oracle with perfect size and difficulty estimates.
\end{enumerate}

Our predictive scheduling framework offers a practical plug-in method for latency- and cost-sensitive LLM deployments, showing that pre-run budget and difficulty prediction can substantially improve inference efficiency without any modifications to the underlying language model.

\section{Related Work}

\subsection{Chain-of-Thought and Meta-Decoding}
Large language models improve multi-step reasoning by generating intermediate “thought” steps. Chain-of-Thought prompting decomposes problems into sub-steps to boost accuracy on arithmetic and logical tasks \citep{wei2022chain}, while self-consistency decoding aggregates multiple sampled chains to reduce variance and error \citep{wang2022self}. Tree-of-Thoughts further explores a tree of partial solutions to guide search \citep{yao2023tree}. These methods yield strong gains but assume a fixed chain length or uniform budget per query, incurring substantial extra computation.

\subsection{Batch Scheduling and Token Reordering}
When serving mixed-difficulty requests, naive FIFO or uniform batching can suffer from head-of-line blocking and wasted tokens on easy queries. The vLLM scheduler predicts remaining tokens to reorder and preempt queries for better throughput \citep{liu2023vllm}, and related pre-scheduling techniques reduce latency in multi-tenant serving \citep{zhang2021efficient}. Dynasor dynamically allocates extra compute to queries deemed “hard” at a fixed checkpoint in the reasoning process \citep{fu2024efficient}. These external schedulers improve average latency but rely on surface-level heuristics or single-checkpoint decisions rather than learned per-query estimates of needed reasoning length.

\subsection{Adaptive Compute Within Transformers}
A parallel line of work adapts internal model computation based on input difficulty. Depth-adaptive Transformers vary the number of layers executed per input \citep{elbayad2020depth}, and Mixture-of-Experts architectures route tokens through a subset of expert sub-modules to save compute \citep{fedus2021switch}. These approaches adjust per-input compute “on the fly” but do not provide pre-run, per-query length estimates or batch-level scheduling under a fixed token budget.

\subsection{Per-Query Token-Budget Prediction and Greedy Allocation}
Estimating each query’s optimal reasoning length in advance can eliminate wasted tokens and avoid under-reasoning. Wu et al.\ analyze how accuracy first increases then plateaus or degrades as chain length grows, demonstrating an optimal CoT length exists per problem \citep{wu2025when}. TALE uses few-shot prompts and small post-trained models to predict the optimal token budget for each query \citep{han2024token}. Fu et al.\ train a length-ordering predictor via learning-to-rank to sort requests by relative size before batching \citep{fu2024efficientllmschedulinglearning}. Li et al.\ introduce a learned early-stopping criterion that decides, during generation, whether additional chains are likely to improve correctness \citep{li2024escapeskyhighcostearlystopping}. Damani et al.\ train lightweight predictors (MLP or LoRA) to estimate the marginal reward of adding more reasoning traces and employ a greedy algorithm to allocate the number of traces under a fixed budget \citep{damani2024learninghardthinkinputadaptive}. While these methods forecast or adapt token usage, they do not leverage the rich internal features of the LLM’s hidden states nor systematically identify which transformer layers carry the strongest predictive signal.

\subsection{Curriculum Learning and Difficulty-Aware Batching}
Curriculum learning proposes to organize training data in an easy-to-hard sequence to accelerate convergence and improve generalization \citep{bengio2009curriculum}. Self-paced learning extends this idea by jointly learning the curriculum and model parameters using an implicit difficulty metric that evolves with training progress \citep{kumar2010self}. Theoretical and empirical studies confirm that well-designed curricula can modify the optimization landscape to yield faster convergence without altering the global optimum \citep{hacohen2019power}. In multi-exit inference architectures, curriculum learning has been employed to improve early-exit classifier accuracy under strict latency constraints by progressively introducing harder examples during training \citep{bakhtiarnia2021improving}.

\subsection{Our Contribution}
In contrast to prior work, we train lightweight predictors on hidden-state features extracted from intermediate transformer layers to forecast, \emph{before any generation}, the per-query reasoning length required to meet a given correctness threshold. We then solve a global batch allocation via a greedy algorithm under a fixed total token budget. To our knowledge, we are the first to systematically analyze which transformer layers yield the most informative signals for such predictions, demonstrating that middle layers provide superior accuracy in reasoning-length estimation.

\section{Methods}

This work was largely inspired by two recent advances in the literature: the adaptive allocation of reasoning budget proposed by \citet{damani2024learninghardthinkinputadaptive} and the dynamic scheduling framework introduced by \citet{fu2024efficient} in their Dynasor system. While \citet{damani2024learninghardthinkinputadaptive} focuses on adapting the number of reasoning traces allocated per query and \citet{fu2024efficient} emphasizes adapting size (the token budget per trace) with fixed reasoning budget cutoff thresholds, our methodology leverages the rich information in the hidden states to dynamically allocate the per-query reasoning budget.

\subsection{Dataset Preprocessing and Experimental Setup}

We conducted our experiments on the GSM8K dataset \citep{cobbe2021trainingverifierssolvemath}, a benchmark of grade school math word problems. The dataset contains 7,450 training examples and 1,294 test examples. Each example consists of a natural language math question and its solution expressed as a series of reasoning steps followed by a final numerical answer.

\subsubsection{Data Processing}
\label{subsubsec:dataprocessing}
For each question in both the training and test sets, we produced a 16-dimensional early-stopping probability vector by first tokenizing the question with DeepSeek’s tokenizer to record the input length \(Q\).  We then generated one hundred independent reasoning traces per question using temperature \(0.7\) and top-\(p\) \(0.95\).  Within each trace we inserted a fixed probe string 
\begin{tcolorbox}[colback=blue!10, colframe=blue!40, arc=0mm, boxrule=0.5pt]
"Oh, I suddenly got the answer to the whole problem, \textbf{Final Answer}\textbackslash n\textbackslash n\textbackslash [ \textbackslash boxed\{"
\end{tcolorbox}
at every 16 tokens up to a maximum of 256 tokens; this probe string forces the model to emit a final answer at that point.  After each probe insertion we extracted the model’s answer and compared it to the ground truth.  Finally, for each probe point (16, 32, …, 256 tokens) we computed the fraction of the one hundred traces that were correct, yielding the early-stopping probability for that token budget.

\subsubsection{Training Data Generation}
\label{subsubsec:difficulty_classification}
Training data for our predictors comprised two components: the input features and the target labels. The input features consisted of the hidden‐state representations produced by the DeepSeek model’s encoder. For each question we extracted the 1536‐dimensional hidden state at the [CLS] token from every transformer layer (layers 1 through 28), yielding 28 feature vectors per example. The target labels comprised two sets of values. First, we recorded early‐stopping probability vectors of length 16, where each entry is the fraction of correct answers observed when forcing an answer at successive probe points (16, 32, …, 256 tokens). Second, we derived difficulty labels—easy, medium, or hard—by computing each question’s correctness probability under a 256‐token per-query reasoning budget, and assigning the bottom 20 percentile to “hard,” the top 20 percentile to “easy” (with thresholds $p_{20}=0.18$ and $p_{80}=0.84$), and all others to “medium.” These features and labels for GSM8K queries formed the train and test sets for both our early‐stopping and difficulty‐classification models.

\subsubsection{Data Splits and Validation}
We maintained the original GSM8K train/test split to ensure comparability with previous work. The final test set of 1,294 examples was used only for final evaluation.
The processed dataset statistics are as follows:
\begin{table}[h]
\centering
\renewcommand{\arraystretch}{1.2}
\begin{tabular}{lrrrr}
\hline
\textbf{Split} & \textbf{Total} & \textbf{Easy} & \textbf{Medium} & \textbf{Hard} \\
\hline
Train & 7,450 & \cellcolor{green!15}1,506 & \cellcolor{yellow!15}4,437 & \cellcolor{orange!15}1,507 \\
Test & 1,294 & \cellcolor{green!15}271 & \cellcolor{yellow!15}760 & \cellcolor{orange!15}263 \\
\hline
\end{tabular}
\caption{Dataset statistics after preprocessing and difficulty stratification}
\label{tab:dataset_stats}
\end{table}

To ensure robustness, we verified that the difficulty distribution in the test set closely matched that of the training set. We also confirmed that the average question length and vocabulary distribution were consistent across splits.

On GSM8K data, we display the early stopping probabilities. Notice that for lower token budgets, the probability that the generated answer will be correct is clustered almost wholly around 0. For higher reasoning budgets, the probabilities that the generated answer will be correct for the given reasoning budget are concentrated near 1. Furthermore, there are some questions for which even the greatest reasoning budgets almost always result in an incorrect answer--we would refer to these as hard/impossible questions. For other questions, even a moderate increase in the reasoning budget leads to significant expected accuracy gains.

\begin{figure}[htbp]
  \centering
  \includegraphics[width=\textwidth]{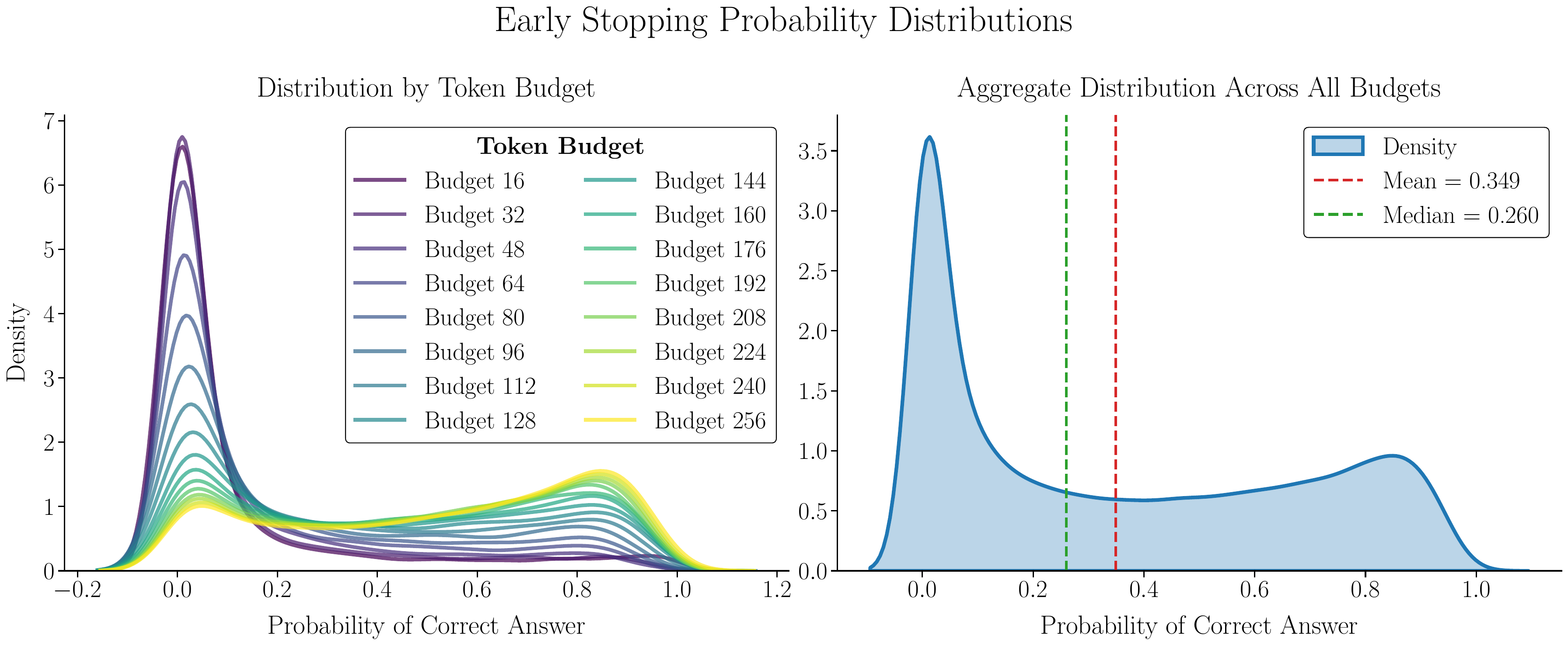}
  \caption{(a) Distribution of correct‐answer probabilities at each token budget; (b) Aggregate KDE over all budgets.}
  \label{fig:cs2241_side_by_side}
\end{figure}

\subsection{Early Stopping Prediction}
\label{subsec:early_stopping_method}
Allocating too few tokens to difficult problems leads to incorrect answers, while allocating too many tokens to simple problems wastes computational resources. By accurately predicting early stopping probabilities—the likelihood that a model can correctly solve a problem after generating a specific number of reasoning tokens—we can make informed decisions about resource allocation. These predictions serve as a foundation for our batch allocation strategies (detailed in Section ~\ref{subsec:greedy_algorithm}), enabling us to dynamically distribute a fixed token budget across diverse queries to maximize overall accuracy.

Inspired by the findings of \citet{damani2024learninghardthinkinputadaptive} that transformer hidden states contain rich signals for predicting an LLM's performance on different queries, we investigate our first research question: \textit{can we predict, before generation, the optimal token budget required on a per-query basis for reasoning tasks to achieve a desired expected accuracy?} 

We explore two complementary approaches to leverage the model's internal representations for predicting early stopping probabilities: (1) lightweight Multi-Layer Perceptrons (MLPs) trained on hidden states from various transformer layers, and (2) a more parameter-efficient LoRA fine-tuning of the base model.

\subsubsection{MLP-based Early Stopping Predictors}

To test our hypothesis that transformer hidden states encode predictive signals about reasoning complexity—and that certain layers may be particularly informative—we train simple MLP models that map hidden state representations to early stopping probability vectors. We specifically hypothesize that middle layers of the transformer architecture, which typically capture a balance of syntactic and semantic features, may provide the strongest predictive signals for reasoning difficulty.

\paragraph{Architecture and Training.} Our MLP predictors follow a consistent architecture comprising two fully connected layers with 256 hidden units and ReLU activation functions, and a final sigmoid activation to constrain outputs to [0,1]. The input dimension corresponds to the hidden state size (1536 for DeepSeek-R1-Distill-Qwen-1.5B), and the output dimension matches the number of prediction points (16, corresponding to token budgets from 16 to 256 in steps of 16). We train separate MLPs for each transformer layer (1-28) to systematically evaluate which layers encode the most predictive signals for reasoning length requirements.

The models are trained using the Adam optimizer and mean squared error (MSE) loss. We employ early stopping based on validation performance to prevent overfitting. To ensure optimal performance for each layer, we conduct a hyperparameter sweep over the number of hidden layers (1 to 3), hidden layer size (128, 256, or 512 units), learning rate ($1 \times 10^{-4}$, $5 \times 10^{-4}$, or $1 \times 10^{-3}$), and dropout rate (0.0, 0.1, or 0.2) to select the best configuration for each layer’s predictor.

\subsubsection{LoRA Fine-tuned Model for Early Stopping Prediction}
As an alternative to extracting and processing hidden states externally,  we hypothesize that early stopping prediction is fundamentally a linguistic task, where semantic understanding of the problem statement correlates with reasoning complexity. Unlike our MLP approach that operates on extracted hidden states, we explore a more integrated approach using Low-Rank Adaptation (LoRA) \citep{hu2022lora}. Our LoRA implementation processes the raw question text as input and directly produces early stopping success probabilities across different token budgets. 

This method enables parameter-efficient fine-tuning by inserting trainable low-rank matrices into the transformer architecture while keeping pre-trained weights frozen. 

Moreover, this design leverages the model's pre-trained linguistic knowledge to identify semantic patterns that we hypothesize correlate with reasoning complexity. For instance, we expect that mathematical questions containing terms like "prove," "derive," or "show that" typically signal higher reasoning complexity requiring longer token budgets, while problems with directives such as "calculate," "find," or "evaluate" often indicate more straightforward computational tasks that can succeed with earlier stopping points. Our LoRA fine-tuning approach aims to capture these linguistic indicators directly from the input text, potentially enabling more accurate early stopping predictions based on the semantic content of the problem statement.

\paragraph{Architecture and Implementation.} Our implementation fine-tunes the DeepSeek-R1-Distill-Qwen-1.5B model directly on the early stopping prediction task. The architecture consists of the base language model augmented with LoRA adapters (rank $r=16$, scaling factor $\alpha=32$) targeting the query and value projection matrices in the attention mechanism. This targeted approach allows us to efficiently adapt the model's reasoning capabilities to our specific task while minimizing parameter count.

\paragraph{Prediction Pipeline.} The prediction pipeline consists of the LoRA-adapted language model followed by a regression head implemented as a two-layer MLP with a hidden dimension of $(h_\text{model}/2)$, where $h_\text{model}$ is the base model's hidden size (1536). We extract the final hidden state corresponding to the last token of the input sequence and process it through layer normalization, dimensionality reduction to $h_\text{model}/2$, ReLU activation, final linear projection to output dimension (16), and sigmoid activation to constrain predictions to $[0,1]$. 

\paragraph{Training Details.} We optimize the model using mean squared error loss between the predicted early‐stopping probabilities and the ground truth. Training is carried out with the AdamW optimizer at a learning rate of $1\times10^{-4}$ for ten epochs, using a batch size of 32 for training and 8 for evaluation. Hyperparameter tuning explores LoRA rank values of 8, 16, or 32; LoRA scaling factors of 16, 32, or 64; learning rates of $5\times10^{-5}$, $1\times10^{-4}$, or $2\times10^{-4}$; and alternative choices for the regression head architecture.

\subsection{Predicting Problem Difficulty for Token Budget Allocation}

While early stopping probability vectors provide fine-grained predictions of reasoning success at various token budgets, they may be unnecessarily detailed for practical allocation strategies. We hypothesize that a simpler, more interpretable approach—classifying problems into discrete difficulty categories—could provide sufficient signal for effective token budget allocation while reducing algorithmic complexity.

The motivation for this classification is twofold: (1) discrete categories map naturally to tiered allocation strategies, where easy, medium, and hard problems receive predetermined token budgets, and (2) classification is potentially more robust to noise than precise probability estimation, especially in deployment scenarios where the distribution of problems may shift from the training data. Furthermore, difficulty classification aligns with human intuition about problem complexity, making the allocation decisions more explainable and trustworthy to users.

We explore two complementary approaches to difficulty classification: a few-shot method leveraging existing capabilities of large language models, and a more specialized LoRA fine-tuned classifier trained specifically for our task.

\subsubsection{Few-Shot Classification with LLMs}

We first investigate whether state-of-the-art LLMs can perform effective few-shot difficulty classification without specialized training. This approach is motivated by the hypothesis that commercially available models have already internalized substantial knowledge about mathematical problem complexity through their pre-training on diverse corpora.

\paragraph{Implementation.}
We implemented a classification pipeline using GPT (o4-mini) to analyze each problem statement and assign a difficulty rating based solely on the question text. The prompt instructed the model to classify the question as easy, medium, or hard, to provide a brief justification for its choice, and to return the result in a structured JSON format suitable for automated processing (see Appendix \ref{sec:fewshot_prompt}).

The prompt defined the difficulty categories using a single labeled example of an easy question, a single labeled example of a medium question, and a single labeled example of a hard question. 

This approach leverages the pre-existing mathematical knowledge and ICL abilities embedded in the LLM to identify linguistic markers of difficulty. For example, the model recognizes that problems containing terms like "calculate the sum" or "find x" typically represent easier tasks, while those involving "prove that" or requiring multi-step reasoning with multiple variables indicate higher difficulty levels. The JSON-formatted responses facilitated automated processing and integration with our token budget allocation system.

\subsubsection{LoRA Fine-tuned Classification Model}

While few-shot classification provides a parameter-efficient baseline, we hypothesize that a specialized classification model fine-tuned on our specific dataset would achieve higher accuracy. To create this specialized classifier, we implemented a LoRA-fine-tuned model derived from the DeepSeek-R1-Distill-Qwen-1.5B base model.

\paragraph{Architecture and Implementation.}
This classification model follows the same input processing as the early‐stopping predictor but produces a discrete label instead of a probability vector. The base language model is extended with LoRA adapters of rank $r=16$ and scaling factor $\alpha=32$, applied to the query and value projection matrices in the attention layers. The classification head begins by applying layer normalization to the final hidden state of the input sequence, then projects it linearly to a dimension of $h_{\text{model}}/2$. A ReLU activation is applied next, followed by dropout with probability 0.1 for regularization. The final step is a linear projection that maps to three output logits corresponding to the easy, medium, and hard classes.

\paragraph{Training Methodology.} We employed cross-entropy loss during training and evaluated performance using accuracy, precision, recall, and F1 score—with particular attention to balanced performance across all difficulty classes. As discussed in Section ~\ref{subsubsec:difficulty_classification}, the ground truth difficulty labels were derived from the 256-token performance data, using the 20th and 80th percentile thresholds from the training set (specifically, $p_{20} = 0.18$ and $p_{80} = 0.84$) to establish the boundaries between easy, medium, and hard problems.

Our implementation uses a batch size of 32 for training and 8 for evaluation, with an AdamW optimizer and learning rate of $1 \times 10^{-4}$. We trained the model for up to 50 epochs with early stopping based on validation accuracy, allowing sufficient time for the model to learn difficulty patterns while preventing overfitting. 

We hypothesize that this fine-tuned classifier will capture more nuanced linguistic patterns specific to our mathematical reasoning tasks than the few-shot approach, while maintaining the interpretability advantages of discrete difficulty categories over continuous probability vectors. The resulting classifications serve as direct inputs to our token budget allocation strategies, enabling differentiated resource allocation based on problem difficulty.

\subsubsection{Greedy Algorithm}
\label{subsec:greedy_algorithm}
The learned MLP or finetuned LORA models are trained on the GSM8K train set, and tested out of sample on the GSM8K test set.

Starting with a minimum allocation of 16 tokens per problem, the algorithm iteratively assigns additional 16-token windows to problems where the predicted gain in accuracy (based on the predicted solution size) is highest. This process continues until either the token budget is exhausted or no further positive gains are expected. 
This approach allows us to adaptively allocate more tokens to problems predicted to require longer solutions while maintaining the average token usage within the specified budget constraint.

\begin{algorithm}[h!]
\caption{Greedy Token Allocation using Predicted Early Stopping Correctness Probabilities}
\begin{algorithmic}[1]
  \Require $Q$, $B$, $W$, $P$  \Comment{queries, budget, window size, probability vectors}
  \Ensure allocations maximizing expected accuracy
  \State allocations $\gets [W]\times|Q|$
  \State remaining $\gets B\cdot|Q| - \sum\text{allocations}$
  \While{remaining $\ge W$}
    \State gains $\gets$ \Call{ComputeGains}{$P,\,$allocations,$\,W$}
    \If{$\max(\mathrm{gains}) \le 0$} \textbf{break} \EndIf
    \State $i^* \gets \arg\max(\mathrm{gains})$
    \State allocations[$i^*$] $\mathrel{+}= W$
    \State remaining $\mathrel{-}= W$
  \EndWhile
  \State \Return allocations
\end{algorithmic}
\vspace{0.5em}
{\footnotesize\textit{Note: ComputeGains returns, for each query, the marginal expected-accuracy gain of adding one more window of size $W$ (or $-\infty$ if no more windows remain).}}
\end{algorithm}

As baselines, we examine the accuracy of the non-adaptive strategy, where the token budget is uniform across all queries, as well as the accuracy of the oracle strategy, where we use the known ground truth early stopping correctness probabilities and allocate tokens using the greedy strategy used above. In order for our proposed method to be useful, it must outperform that nonadaptive baseline. The oracle method provides a notion of the best possible accuracy performance under perfect knowledge of early stopping probabilities, assigning per-query token budgets using the proposed greedy method. 

\begin{figure}[htbp]
  \centering
  \includegraphics[width=\textwidth]{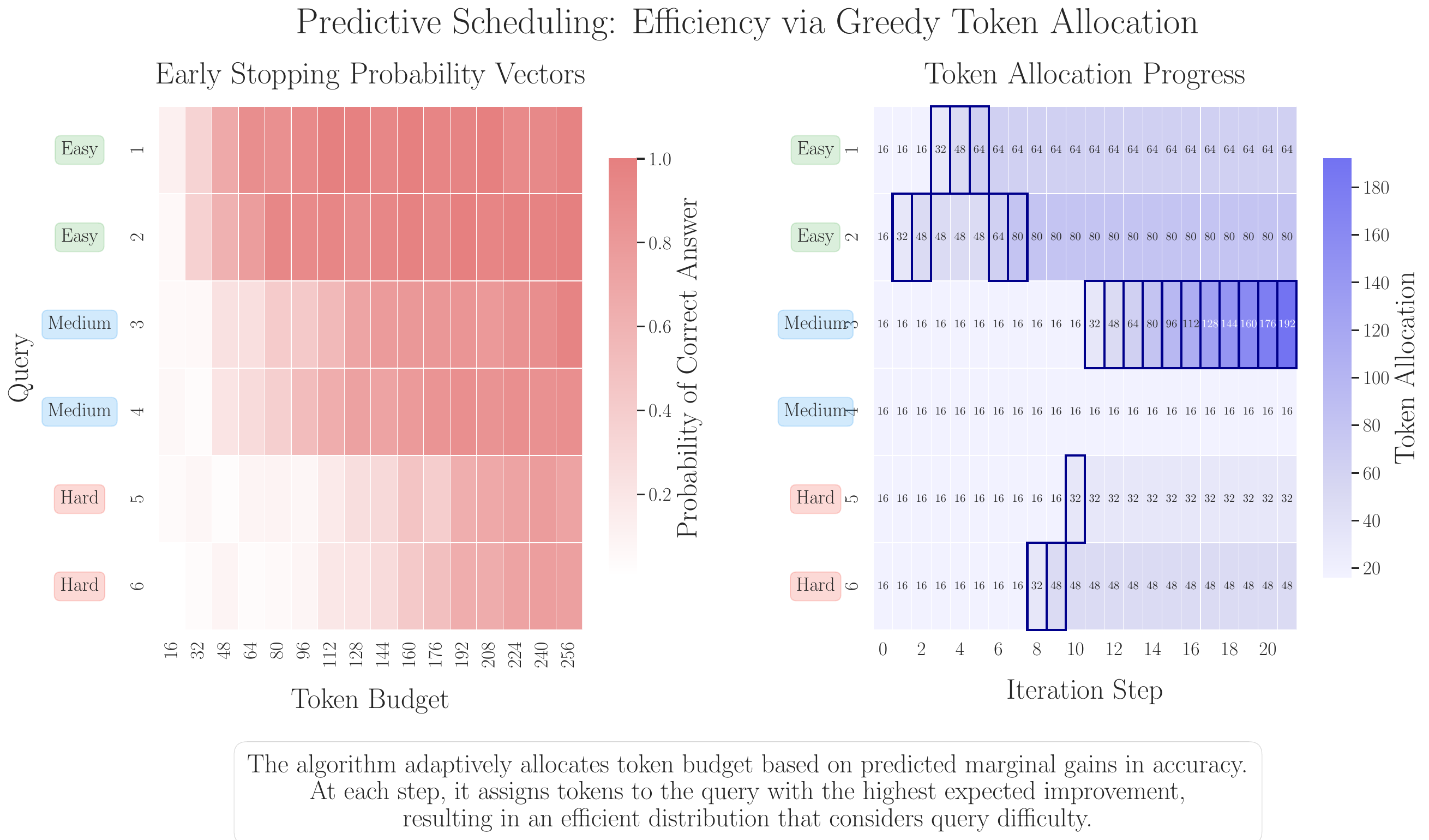}
  \caption{Visualization of the greedy token allocation algorithm. \textbf{Left:} Early stopping probability vectors showing the likelihood of generating a correct answer given different token budgets for each query. Darker shades indicate higher probability of correctness. \textbf{Right:} Progressive token allocation across queries during algorithm execution. Each cell shows the allocated token budget, and blue outlines highlight the query receiving additional tokens at each step. The algorithm initially allocates a minimum budget to all queries, then iteratively assigns additional tokens to queries with the highest expected marginal accuracy gain, prioritizing easier queries at first but gradually shifting resources to more difficult ones as the budget increases.}
  \label{fig:greedy-allocation}
\end{figure}

\subsection{Difficulty-Based Token Budget Allocation}
We partition the inference-time queries into three classes—easy, medium, and hard—using the difficulty predictions described in Section~\ref{subsubsec:difficulty_classification}.  Let $p_1$, $p_2$, and $p_3$ denote the fractions of queries in the current inference batch predicted as easy, medium, and hard, respectively.  For each category $k\in\{\mathrm{easy},\mathrm{medium},\mathrm{hard}\}$, define
\[
  \mathrm{acc}_{k}(b)
  = \frac{1}{N_k}\sum_{q_i\in \mathcal{C}_k} \Pr\bigl(\text{correct answer for }q_i\mid b\text{ reasoning tokens}\bigr),
\]
where $\mathcal{C}_k$ is the set of all training examples in category $k$ and $N_k=|\mathcal{C}_k|$.  Thus $\mathrm{acc}_{k}(b)$ is the expected accuracy when allocating $b$ tokens per query to problems of difficulty $k$, computed from the element-wise average of all early-stopping probability vectors of problems of difficulty $k$ in the GSM8K training set.

Under an average per-query budget $B$ and a window size $W$, we search for the optimal per-category budgets $(b_1,b_2,b_3)\in\{W,2W,\dots,16W\}^3$ by solving
\[
  \max_{b_1,b_2,b_3}\;p_1\,\mathrm{acc}_{\mathrm{easy}}(b_1)
  + p_2\,\mathrm{acc}_{\mathrm{medium}}(b_2)
  + p_3\,\mathrm{acc}_{\mathrm{hard}}(b_3)
  \quad\text{s.t.}\quad p_1b_1 + p_2b_2 + p_3b_3 \le B.
\]
The inputs to Algorithm~2 are the set of inference queries $Q$, the budget $B$, the window size $W$, the vector of difficulty predictions, and the batch proportions $p_1,p_2,p_3$.  After computing $(b_1,b_2,b_3)$, each query in class $k$ receives $b_k$ reasoning tokens, as shown below.

\begin{algorithm}[htbp]
\caption{Difficulty-Based Token Budget Allocation}
\begin{algorithmic}[1]
  \Require $Q,\,B,\,W,\,\mathit{difficulty\_predictions},\,p_1,p_2,p_3$
  \Ensure allocations per query maximizing expected accuracy
  \State $(b_1,b_2,b_3)\gets \Call{GetOptimalBudgets}{B,p_1,p_2,p_3}$
  \For{$i=1,\dots,|Q|$}
    \If{difficulty\_predictions[$i$]=\texttt{easy}}
      \State allocations[$i$] $\gets b_1$
    \ElsIf{difficulty\_predictions[$i$]=\texttt{medium}}
      \State allocations[$i$] $\gets b_2$
    \Else
      \State allocations[$i$] $\gets b_3$
    \EndIf
  \EndFor
  \State \Return allocations
\end{algorithmic}
\vspace{0.5em}
{\footnotesize\textit{Note: \textsc{GetOptimalBudgets} exhaustively searches all triples in $\{W,2W,\dots,16W\}^3$ and returns the one maximizing $p_1\,\mathrm{acc}_{\mathrm{easy}}(b_1) + p_2\,\mathrm{acc}_{\mathrm{medium}}(b_2) + p_3\,\mathrm{acc}_{\mathrm{hard}}(b_3)$ subject to $p_1b_1 + p_2b_2 + p_3b_3 \le B$.}}
\end{algorithm}
\noindent This allocation strategy tailors the reasoning budget per difficulty class using both the train-set accuracy curves and the predicted class proportions in the inference time batch, yielding superior performance over uniform budgeting (see Section~\ref{sec:results}).

\section{Results}
\label{sec:results}

\subsection{Early Stopping Prediction Results}

We evaluate our two approaches for predicting early stopping probabilities: (1) MLPs trained on hidden state features from different transformer layers, and (2) a LoRA fine-tuned model operating directly on question text. Our analysis specifically addresses the two key hypotheses we formulated in Section \ref{subsec:early_stopping_method}: whether middle transformer layers encode stronger predictive signals about reasoning complexity, and whether linguistic features in problem statements correlate with required reasoning length.

\subsubsection{Layer-wise Analysis of Hidden State Features}

Our first hypothesis posited that specific layers of the transformer architecture—particularly middle layers—would encode the strongest predictive signals about reasoning complexity. To test this hypothesis, we trained identical MLPs on hidden states extracted from each of the 28 layers of DeepSeek-R1-Distill-Qwen-1.5B, with each MLP predicting the probability of correct answers at different reasoning budgets.

\begin{figure}[h!]
\centering
\includegraphics[width=\textwidth]{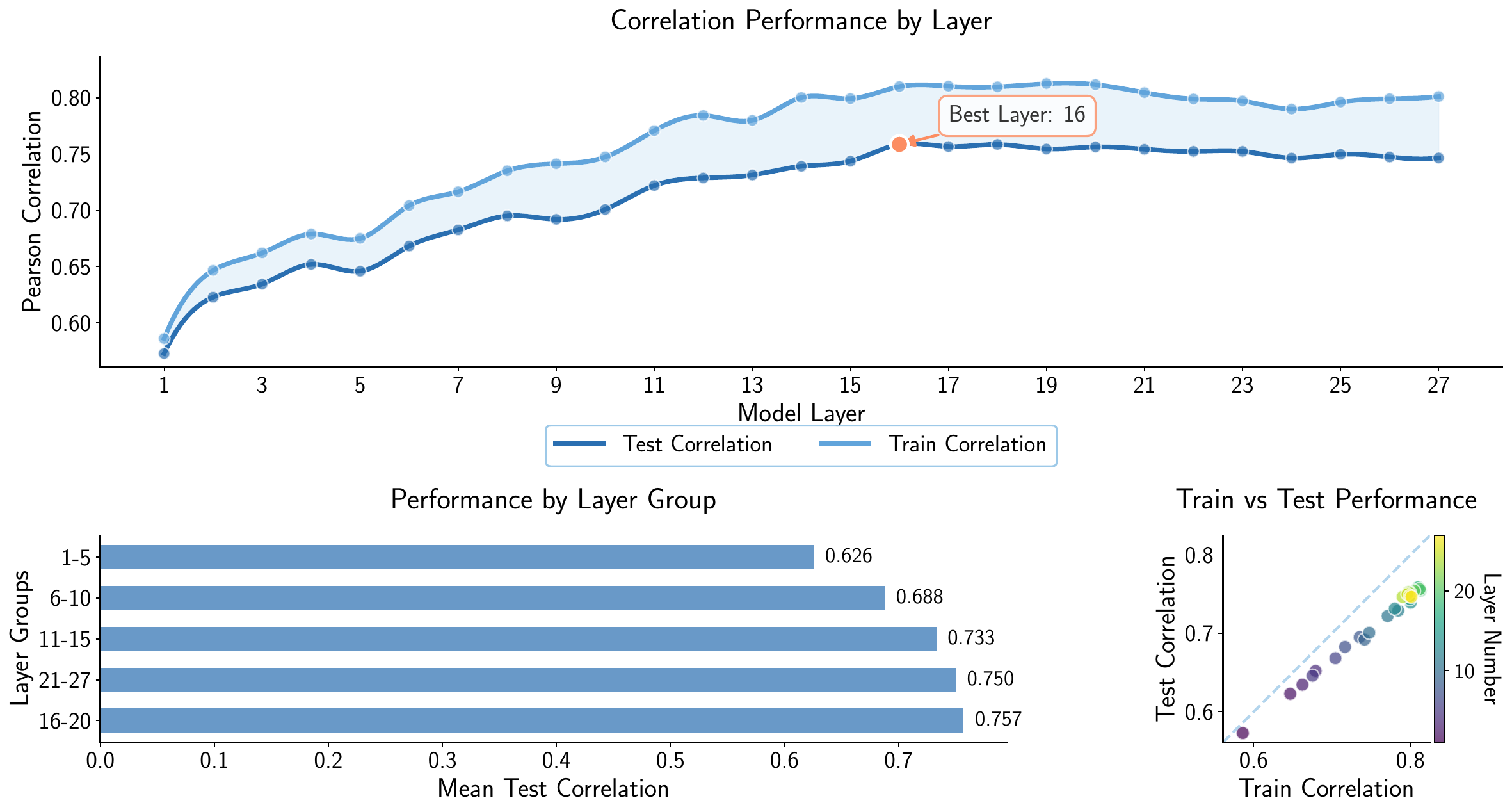}
\caption{Correlation performance analysis across model layers. The top panel displays the Pearson correlation coefficients for both test and train datasets achieved by MLPs trained on different layer features of the DeepSeek-R1-Distill-Qwen-1.5B model. The middle layers (particularly layer 16) achieve the highest correlation for predicting early stopping performance, suggesting that intermediate representations offer the strongest signal for reasoning difficulty prediction. The bottom left panel shows aggregated performance by layer group. The bottom right panel illustrates the relationship between train and test correlation, with points colored by layer number.}
\label{fig:correlation}
\end{figure}

Figure~\ref{fig:correlation} presents comprehensive evidence supporting our hypothesis. Middle layers (particularly 12-17) significantly outperform both early and late layers, with layer 16 achieving the highest test correlation of 0.742. This confirms our prediction that intermediate representations capture an optimal balance of syntactic and semantic features relevant to reasoning difficulty.

The performance distribution follows a clear inverted U-shape across model depth, with early layers (1-6) showing correlations below 0.6 and late layers (21-28) dropping to similar levels. This pattern aligns with our hypothesis that early layers primarily capture surface-level features insufficient for reasoning complexity assessment, while later layers become too specialized toward output generation to retain general reasoning signals.

\begin{figure}[h!]
\centering
\includegraphics[width=\textwidth]{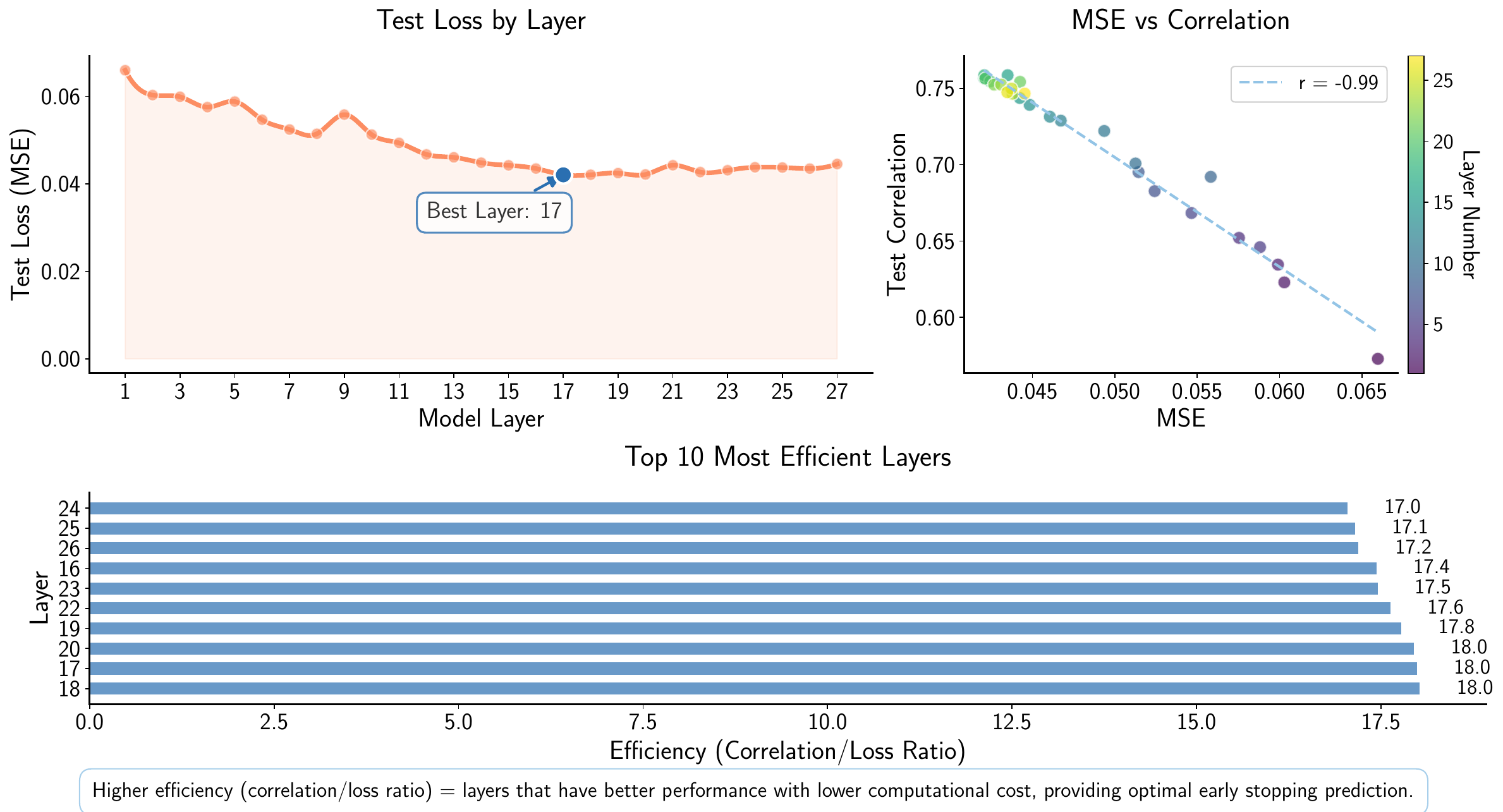}
\caption{The top left panel shows test loss (MSE) by layer, demonstrating how prediction error varies across the model's depth. The top right panel illustrates the relationship between MSE and Pearson correlation, revealing that while these metrics are generally inversely related, some layers achieve better correlation despite similar loss values. The bottom panel presents correlation-to-loss ratio for the top performing layers, indicating which layers provide the most predictive value per unit of loss.}
\label{fig:loss}
\end{figure}

Figure~\ref{fig:loss} examines prediction efficiency using the correlation-to-loss ratio, which measures predictive power per unit of error. This analysis further confirms the effectiveness of middle layers (12-17), with these layers achieving 15-20\% higher efficiency than early or late layers. This quantitative result strongly supports our hypothesis about the advantageous position of middle transformer layers for reasoning complexity prediction.

While our results validate the core hypothesis that hidden states—particularly from middle layers—encode significant information about reasoning complexity, the maximum correlation of 0.742 suggests this relationship is more complex than initially theorized. The moderate correlation indicates that while transformer hidden states contain substantial predictive signals, additional factors not captured in these representations likely influence reasoning difficulty as well. This partial confirmation suggests that our hypothesis about hidden state informativeness was correct in direction but perhaps optimistic in magnitude.

\subsubsection{Linguistic Features and Reasoning Complexity}

Our second hypothesis proposed that linguistic patterns in problem statements directly correlate with reasoning complexity, and that a LoRA fine-tuned model could capture these patterns to predict early stopping probabilities. This approach achieved an evaluation MSE of 0.0795 and a Pearson correlation coefficient of 0.444 between predicted and ground truth probabilities.

The moderate positive correlation of 0.444 indicates that linguistic features in problem statements do contain signals about reasoning complexity, providing partial support for our hypothesis. However, the correlation strength falls short of the MLP approach, suggesting that predicting exact early stopping probabilities—a continuous, fine-grained task—from linguistic features alone is more challenging than we initially hypothesized.

 While linguistic features appear insufficient for precise prediction of continuous early stopping probability vectors, they may still effectively predict discrete difficulty categories (as we show in our difficulty classification results in Section~\ref{subsubsec:difficulty_classification}). This indicates that linguistic patterns in problem statements may better correlate with coarse-grained difficulty assessment than with fine-grained reasoning length prediction.

The LoRA-based predictor demonstrated lower performance compared to our best MLP model (correlation of 0.444 vs. 0.742) for continuous early stopping prediction. Several factors may explain this performance difference: First, predicting a 16-dimensional vector of probabilities requires more precise mapping than discrete classification, potentially exceeding what can be effectively learned from linguistic features alone. Second, hidden states may represent already-transformed features that more directly correlate with reasoning performance, whereas raw linguistic features require substantial additional processing to yield similar predictive power. Third, the limited scope of LoRA adaptation may be insufficient for the complex mapping from linguistic features to precise early stopping probabilities, though it may be adequate for coarser difficulty classification.

While showing lower performance for continuous probability prediction, the LoRA approach still offers valuable complementary insights. Its moderate positive correlation confirms that linguistic signals do relate to reasoning complexity, even if additional transformative processing (as occurs in middle transformer layers) enhances these signals. Additionally, the end-to-end nature of this approach offers practical advantages for deployment scenarios where access to internal model states may be limited.

Our results support our first hypothesis regarding the advantageous status of middle transformer layers for reasoning complexity prediction. Our second hypothesis about linguistic features and reasoning complexity requires refinement—while linguistic features do correlate with reasoning complexity, they appear better suited for coarse-grained difficulty classification than precise early stopping prediction. As we will show in Section~\ref{subsubsec:difficulty_classification}, the same linguistic approach achieves substantially higher performance when applied to discrete difficulty classification rather than continuous probability prediction.

\begin{figure}[htbp]
\centering
\includegraphics[width=\textwidth]{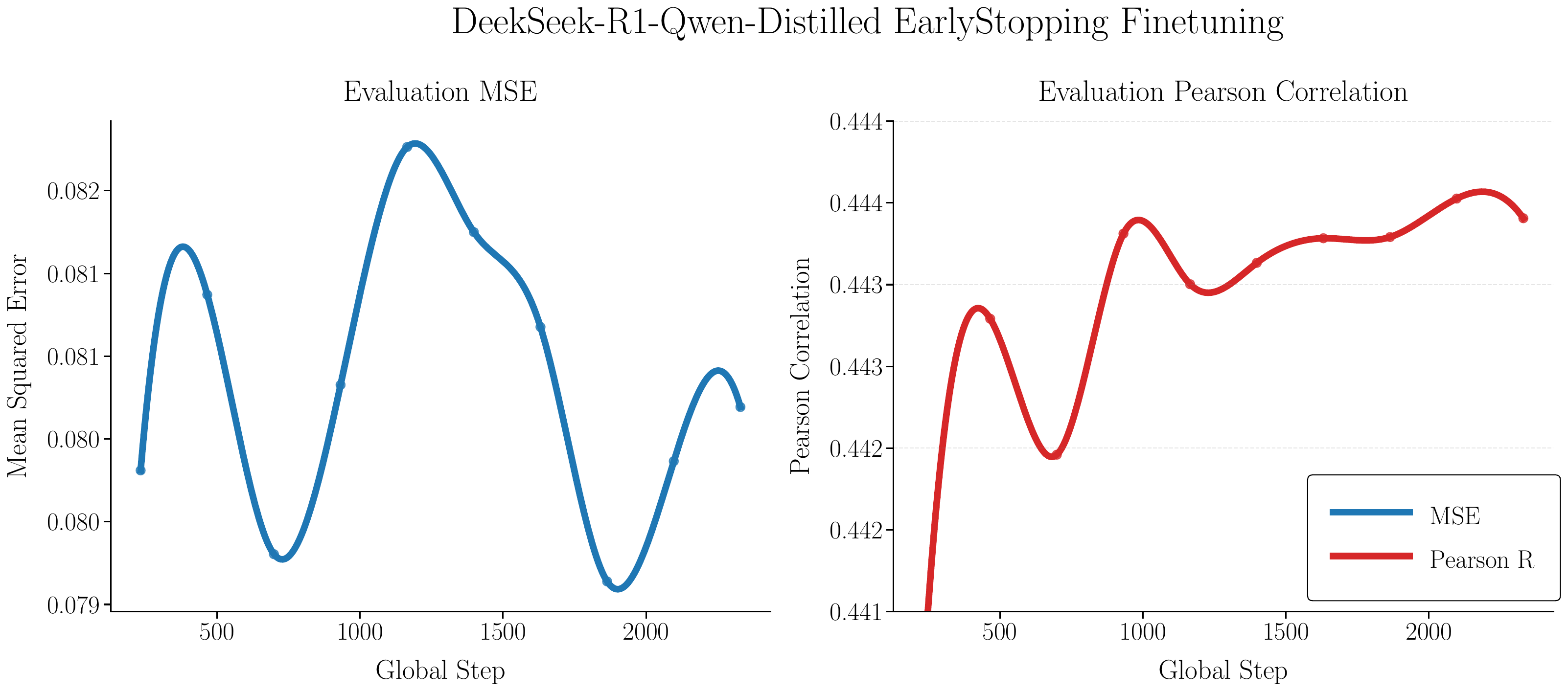}
\caption{Training DeepSeek-R1-Qwen-Distilled model with LoRA fine-tuning for early stopping prediction. \textbf{Left:} Evaluation MSE \textbf{Right:} Evaluation Pearson correlation coefficient between predicted and ground truth early stopping probabilities.}
\label{fig:lora_finetuning}
\end{figure}

\subsubsection{Token Budget Allocation Using Early Stopping Predictions}

Having established that transformer hidden states—particularly from middle layers—encode meaningful signals about reasoning complexity, we investigate whether these predictions can effectively guide token budget allocation in practical deployment scenarios. Our goal is to determine if adaptive allocation strategies based on early stopping predictions can outperform uniform budget distribution across queries.

We implemented a greedy allocation algorithm that leverages our MLP predictions from layer 16 (our best-performing layer) to distribute a fixed token budget across a batch of problems. The algorithm first assigns a minimum budget to all problems, then iteratively allocates additional tokens to problems predicted to gain the most accuracy from increased reasoning budget.

As baselines, we examine the accuracy of the non-adaptive strategy, where the token budget is uniform across all queries, as well as the accuracy of the oracle strategy, where we use the known early stopping correctness probabilities and allocate tokens using the same greedy strategy. In order for our proposed method to be useful, it must outperform the non-adaptive baseline. The oracle method provides a notion of the best possible accuracy performance given perfect predictive abilities using the greedy method.

\begin{figure}[htbp]
  \centering
  \includegraphics[width=1.0\linewidth]{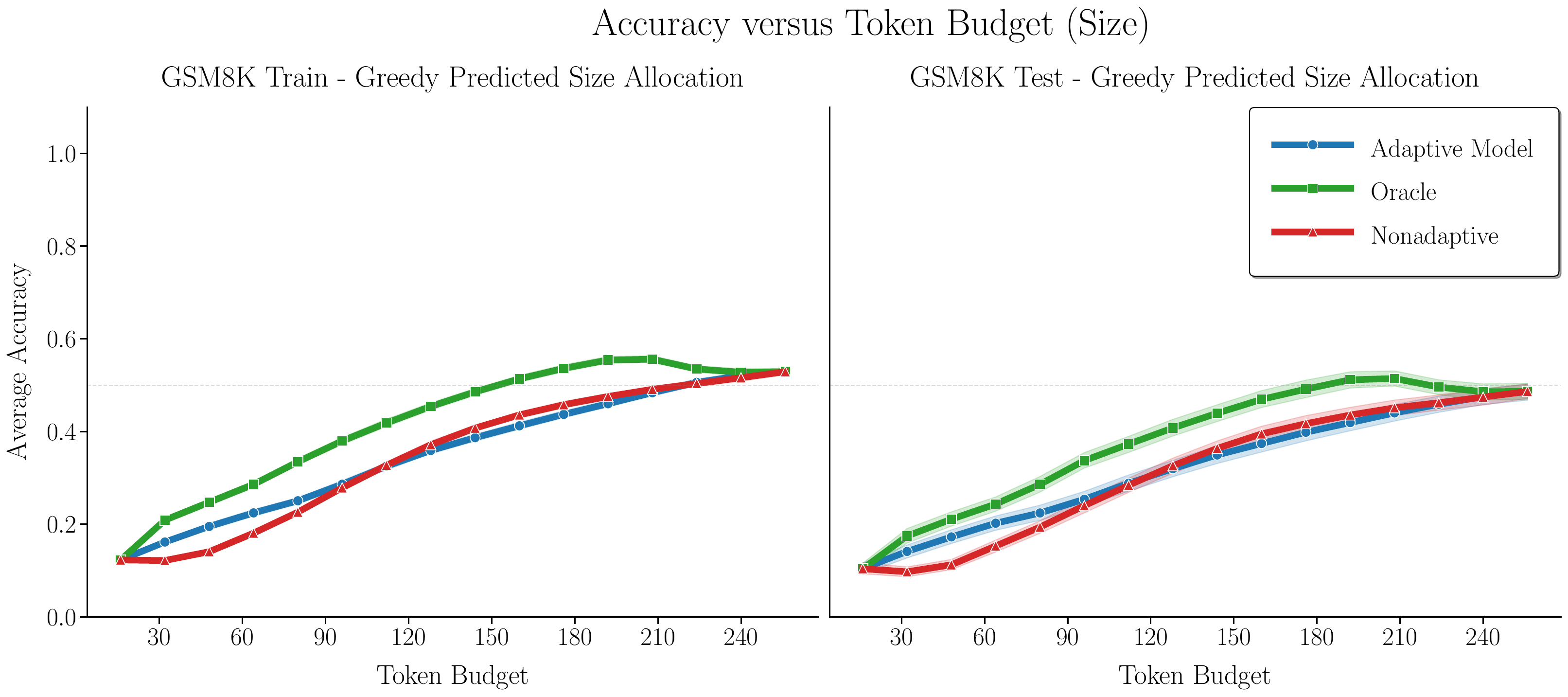}
  \caption{Accuracy vs.\ Token Budget for adaptive allocation using early stopping predictions from a 2-layer MLP trained on layer 16 hidden states. The plot compares three strategies: (1) our adaptive allocation based on MLP predictions, (2) a non-adaptive baseline with uniform allocation, and (3) an oracle allocation using ground truth early stopping probabilities. The adaptive approach outperforms uniform allocation in the constrained budget regime (16-96 tokens per query) but falls below the baseline at higher token budgets, suggesting that prediction errors become more consequential as budget constraints relax.}
  \label{fig:cs2241-size-accuracy}
\end{figure}

Figure~\ref{fig:cs2241-size-accuracy} presents the comparative performance of these allocation strategies across different token budgets. For smaller token budgets between 16-96 reasoning tokens on average per query, the adaptive allocation method using MLP model predictions outperforms the non-adaptive method. This improvement is most pronounced in the most constrained budget regimes (16-48 tokens).

For higher average token budgets, however, the model predictions prove insufficiently accurate, and the adaptive model underperforms the non-adaptive baseline. This performance inversion occurs around 96-128 tokens per query, suggesting a crossover point where the cost of prediction errors begins to outweigh the benefits of adaptive allocation. As budget constraints relax, the uniform allocation strategy eventually allocates sufficient tokens to most problems, reducing the advantages of adaptivity.

The gap between our adaptive allocation and the oracle performance indicates substantial room for improvement in early stopping predictions. This gap is particularly pronounced at higher token budgets, suggesting that while our current predictors capture enough signal for effective allocation in constrained settings, they struggle to identify the optimal stopping points for problems requiring longer reasoning chains.

\subsection{Discrete Difficulty Classification Results}

Building upon our early stopping prediction work, we investigate the effectiveness of classifying problems into discrete difficulty categories (easy, medium, hard) for token budget allocation. This section presents our findings on: (1) the validity of using averaged early stopping vectors for difficulty-based categorization, (2) comparative performance of few-shot versus fine-tuned difficulty classifiers, and (3) effectiveness of difficulty-based token allocation.

\subsubsection{Validation of Difficulty-Based Categorization}

Before implementing difficulty-based allocation strategies, we needed to verify that our discrete categorization meaningfully captures differences in early stopping behavior. Specifically, we investigated whether element-wise averaging of early stopping probability vectors within each difficulty class provides a reasonable basis for allocation decisions.

\begin{figure}[htbp]
  \centering
  \includegraphics[width=1.0\linewidth]{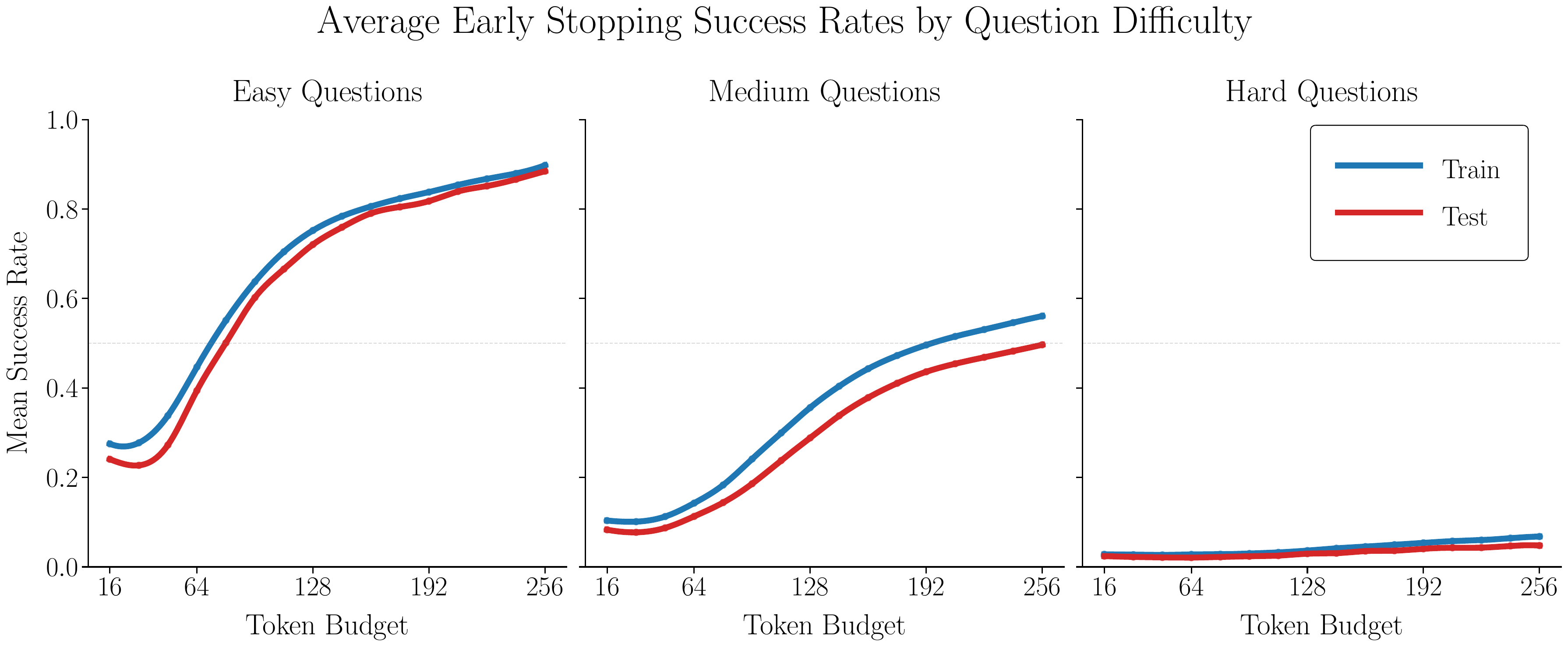}
  \caption{Early stopping probability vectors averaged across problems in each difficulty category (easy, medium, hard from left to right). The x-axis represents token budgets from 16 to 256 tokens in 16-token increments, while the y-axis shows the probability of generating a correct answer. Blue lines represent training set averages, and red lines represent test set averages. Note that the patterns are consistent between train and test sets within each difficulty category, validating our categorization approach. Easy problems show rapid improvement with additional tokens but plateau earlier, medium problems show more gradual improvement across the token range, and hard problems maintain consistently low success probabilities until the highest token budgets.}
  \label{fig:early_stopping_by_difficulty}
\end{figure}

Figure~\ref{fig:early_stopping_by_difficulty} presents a comparative analysis of early stopping probability curves across the three difficulty categories. The plots reveal distinct patterns: easy problems (left panel) show rapid improvement in success probability with increasing token budget, medium problems (center panel) demonstrate more gradual improvement, and hard problems (right panel) maintain consistently low success probabilities across most token budgets.

The averaged vectors for train and test sets (blue and red lines, respectively) track closely within each difficulty category, indicating that these categorizations capture consistent patterns that generalize across data splits. The hard category shows the most consistent behavior across different token budgets, with success probabilities remaining near zero for most budget levels before showing minimal improvement at the highest budgets. In contrast, the easy category exhibits the greatest variability, with success probabilities rising sharply between 16 and 96 tokens before plateauing.

This analysis confirms that our tripartite difficulty categorization captures fundamentally different reasoning patterns that persist across data splits. The consistency between train and test curves validates our approach of using averaged early stopping vectors as the basis for difficulty-based token allocation strategies.

\subsubsection{Comparative Analysis of Difficulty Classification Models}

To test our hypothesis that linguistic features in problem statements correlate with reasoning complexity, we implemented and evaluated two approaches for difficulty classification: (1) few-shot classification using the o4-mini-high model and (2) LoRA fine-tuning of a DeepSeek-1.5B model.

The results provide substantial evidence supporting our linguistic complexity hypothesis. The LoRA fine-tuned model significantly outperformed the few-shot approach, achieving 66.3\% test accuracy compared to 41.6\% for few-shot classification. This 24.7 percentage point improvement demonstrates that linguistic patterns in problem statements contain sufficient signal for meaningful difficulty classification.

\begin{figure}[h!]
  \centering
  \includegraphics[width=1.0\linewidth]{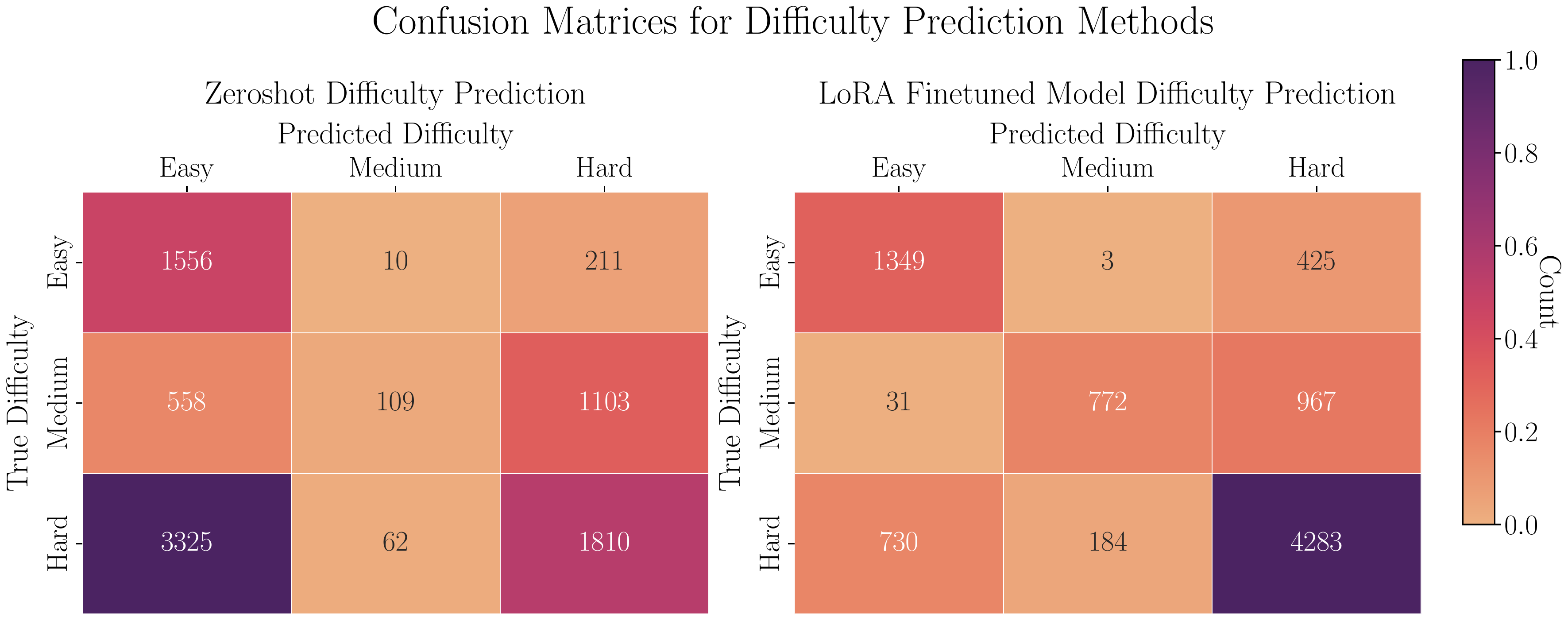}
  \caption{Confusion matrix of predicted vs actual difficulties for both the few-shot and LoRA finetuned prediction models. The matrices reveal that both models struggle most with the medium category, though the LoRA finetuned model (right) shows substantially stronger diagonal elements (66.3\% overall accuracy) compared to the few-shot approach (left, 41.6\% accuracy). }
  \label{fig:cs2241_confusion}
\end{figure}

Figure~\ref{fig:cs2241_confusion} presents confusion matrices for both classification approaches. The LoRA fine-tuned model shows stronger diagonal elements, indicating better classification across all difficulty levels. The performance pattern aligns with our hypothesis that problem statements contain linguistic markers of difficulty—terms like "calculate" or "find" for easier problems versus "prove" or multi-step reasoning indicators for harder problems. Notably, both models struggle most with the medium category, frequently misclassifying these problems as either easy or hard. This confusion likely stems from the inherent ambiguity at category boundaries, as our difficulty categorization imposes discrete labels on what is fundamentally a continuous spectrum of reasoning complexity.

These results, when compared with our earlier early stopping prediction findings, provide important nuance to our earlier linguistic complexity hypothesis. While linguistic features proved insufficient for precise prediction of continuous early stopping probability vectors (achieving correlation of only 0.444), they demonstrate much stronger predictive power for discrete difficulty classification (66.3\% accuracy). This contrast suggests a refinement of our original hypothesis: linguistic patterns in problem statements do correlate with reasoning complexity, but this relationship is more effectively captured through coarse-grained categorization than through fine-grained continuous prediction.

The successful application of linguistic features for difficulty classification, despite their limitations for detailed early stopping prediction, indicates that our hypothesis about linguistic markers of reasoning complexity was correct in principle but required specification about the appropriate granularity of prediction. 

\subsubsection{Difficulty-Based Token Allocation Performance}

Leveraging our difficulty classifiers, we implemented a greedy token allocation algorithm that distributes a fixed token budget across problems based on their predicted difficulty levels. The algorithm begins by assigning each problem a minimum allocation of 16 tokens, then iteratively allocates additional 16-token windows based on expected accuracy gains for each difficulty class.

This allocation strategy relies on pre-computed mappings from difficulty levels to expected accuracy improvements per token window, derived from the averaged early stopping vectors shown in Figure~\ref{fig:early_stopping_by_difficulty}. At each iteration, the algorithm prioritizes allocating tokens to problems where the predicted marginal accuracy gain is highest, continuing until either the token budget is exhausted or no further positive gains are expected.

\begin{figure}[h!]
  \centering
  \includegraphics[width=1.0\linewidth]{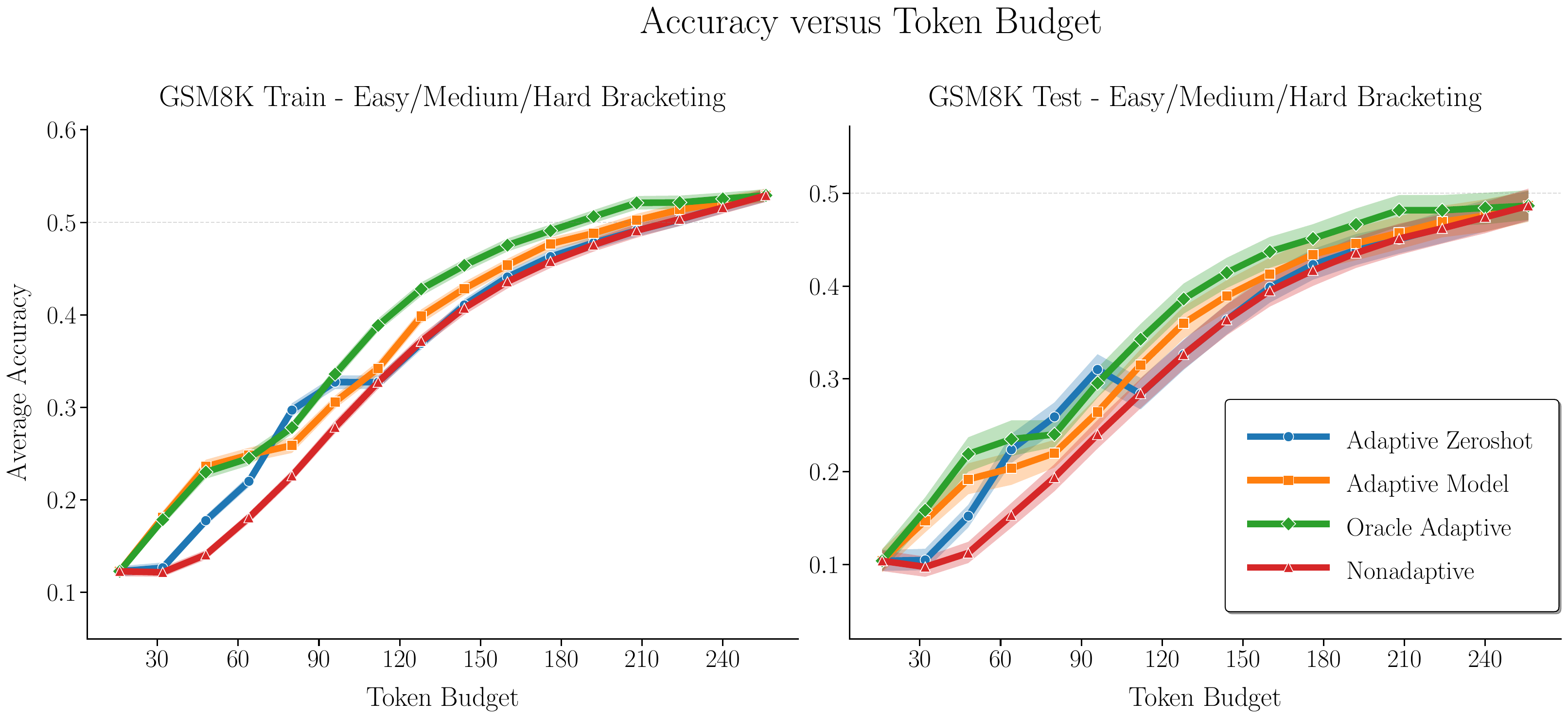}
  \caption{Accuracy vs.\ Token Budget using few-shot or LoRA-finetuned difficulty predictions for token allocation. The plot demonstrates that LoRA-finetuned difficulty classification consistently leads to more effective token allocation across all budget levels, with the performance advantage becoming particularly pronounced at intermediate token budgets (80-160 tokens per problem).}
  \label{fig:cs2241-difficulty-accuracy}
\end{figure}

Figure~\ref{fig:cs2241-difficulty-accuracy} presents the accuracy-versus-token-budget curves for both few-shot and LoRA-based difficulty prediction. The LoRA-based approach outperforms few-shot classification across most token budgets, with the performance gap widening at intermediate budgets (approximately 80-160 tokens per problem).

\begin{figure}[h!]
  \centering
  \includegraphics[width=0.8\linewidth]{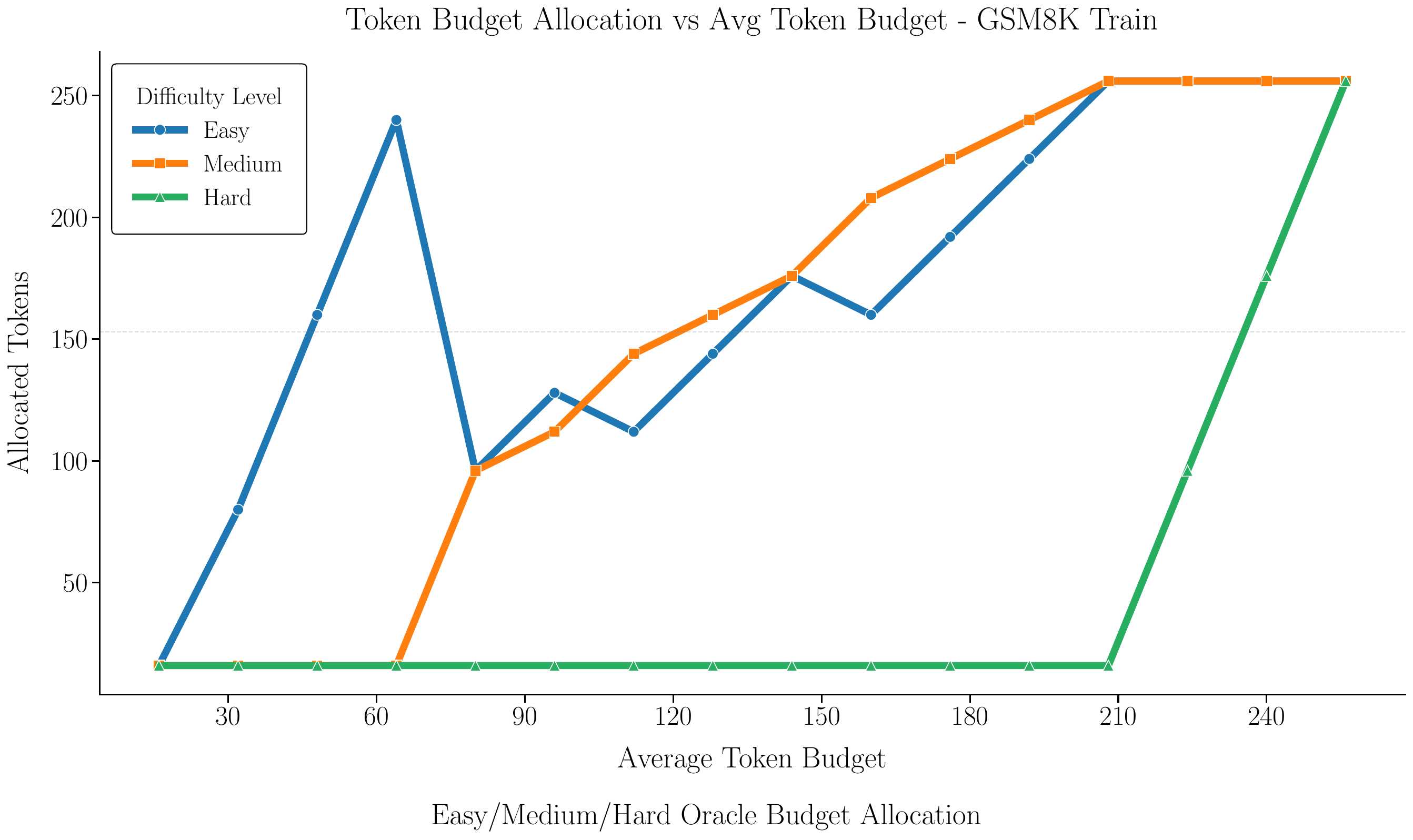}
  \caption{For lower token budgets, in the oracle greedy algorithm, most reasoning budget is allocated to easy questions to maximize accuracy gains. For increasing token budget, more tokens are allocated to medium questions and then last to hard questions to maximize expected accuracy gains. This allocation pattern emerges naturally from the different expected accuracy trajectories of each difficulty category, where easy problems offer high initial returns but quick saturation, while hard problems require substantial investment before showing meaningful improvement.}
  \label{fig:cs2241-difficulty-allocation}
\end{figure}

The allocation patterns themselves reveal insights about optimal resource distribution. Figure~\ref{fig:cs2241-difficulty-allocation} shows that at constrained token budgets, the algorithm heavily favors easy problems, which offer the highest marginal accuracy gains per token. As the budget increases, allocation gradually shifts toward medium problems and finally to hard problems at the highest budgets. This behavior emerges naturally from the expected accuracy curves—easy problems show steep initial gains but quick saturation, while hard problems require substantial token investment before showing meaningful improvements.

Comparing across allocation strategies, our difficulty-based approach proves particularly effective at intermediate token budgets, where the constraints force non-trivial allocation decisions. In these regimes, difficulty-based allocation outperforms uniform allocation by identifying which problems would benefit most from additional computation. Interestingly, despite using coarser categorization than our continuous early stopping predictors, the difficulty-based allocation achieves comparable or better performance in many budget regimes, suggesting that the simplification to discrete categories preserves most of the signal needed for effective allocation while being more robust to prediction noise.

The effective performance of difficulty-based allocation, particularly with the LoRA-based classifier, provides strong support for our refined linguistic hypothesis. While predicting precise early stopping probabilities from linguistic features proved challenging, classifying problems into difficulty categories based on those same features is highly effective. This finding suggests that linguistic patterns in problem statements do indeed correlate with reasoning complexity, but this relationship is more robustly captured through discrete categorization than continuous vector prediction.

\subsection{Size vs. Difficulty: Which Prediction Framework Yields Better Allocation?}

After evaluating both early stopping prediction and difficulty classification approaches individually, a crucial question remains: \textit{which framework provides more effective guidance for token budget allocation in practical deployment scenarios?} To address this question directly, we conducted a comparative analysis of allocation strategies based on both approaches under identical token budget constraints.

\begin{figure}[htbp]
 \centering
 \includegraphics[width=1\linewidth]{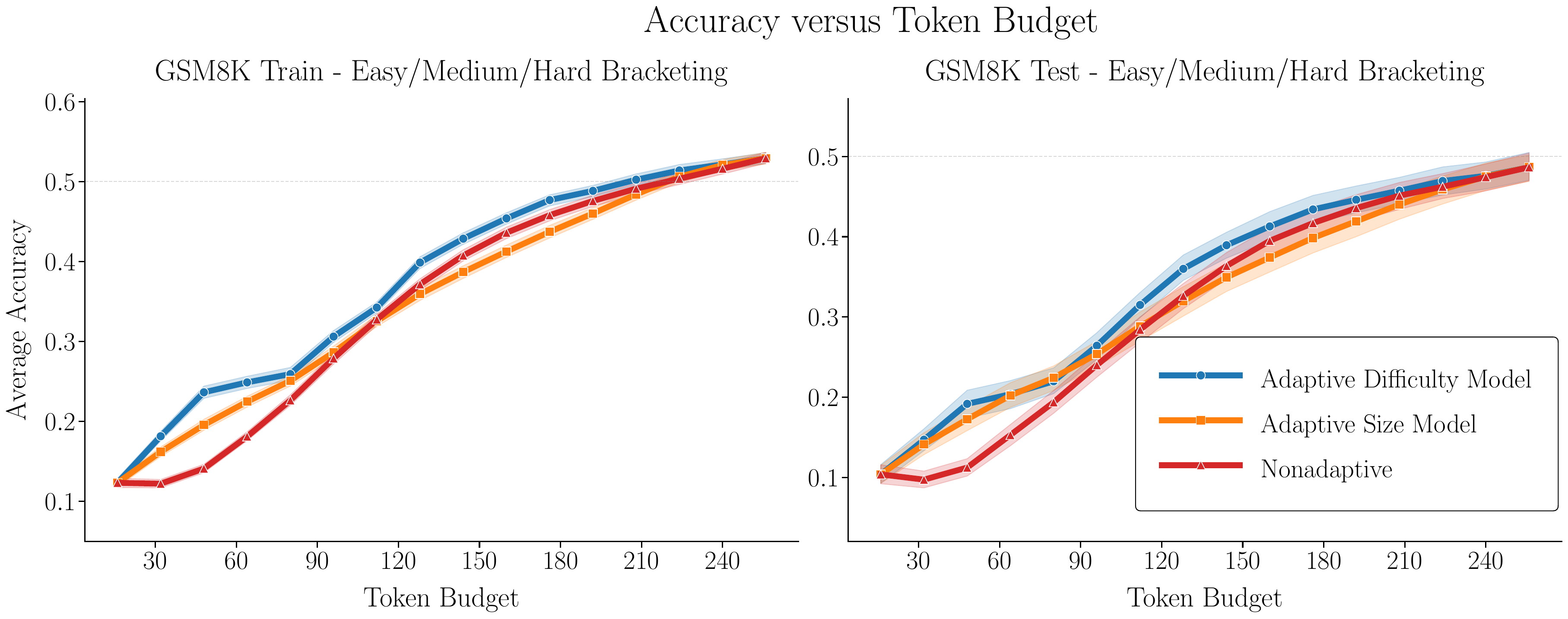}
 \caption{Comparative analysis of allocation strategies across fixed per-query token budgets. The figure shows accuracy achieved by different allocation methods: difficulty-based allocation (using LoRA fine-tuned predictions), size-based allocation (using MLP layer 16 predictions), and the non-adaptive baseline (uniform allocation). Results demonstrate that dynamically allocating based on difficulty predictions consistently outperforms size-based allocation on both train and test sets across all evaluated token budgets, while both adaptive methods outperform the non-adaptive baseline in most budget regimes.}
 \label{fig:cs2241-accuracy-comparison}
\end{figure}

Figure~\ref{fig:cs2241-accuracy-comparison} presents a side-by-side comparison of accuracy achieved by different allocation methods across fixed per-query token budgets. The results reveal a consistent pattern: difficulty-based allocation outperforms size-based allocation across all evaluated token budgets for both train and test sets. This performance advantage persists despite the seemingly coarser granularity of the tripartite difficulty classification compared to the continuous early stopping probability vectors.

Several factors may explain the surprising effectiveness of difficulty-based allocation. First, the discrete categorization provides robustness against prediction noise, as small errors in early stopping probability estimates can significantly impact allocation decisions, while misclassifications between adjacent difficulty categories often result in only minor allocation adjustments. Second, the difficulty-based approach captures qualitatively different reasoning patterns that may not be fully represented in the continuous early stopping curves, such as problems that show sudden accuracy jumps at specific token thresholds. Third, the allocation algorithm based on difficulty categories can leverage more stable, category-level statistics about expected accuracy improvements, reducing vulnerability to outlier patterns.

\section{Conclusion}

We have presented a framework for inference-time reasoning in large language models that leverages fast, pre-run predictors to estimate each query's optimal reasoning-trace length or difficulty category. Our systematic investigation revealed several key insights: (1) hidden-state features from transformer layers contain significant predictive information about reasoning complexity, with middle layers (12-17) providing the strongest signals; (2) linguistic features in problem statements correlate with reasoning difficulty, but this relationship is more effectively captured through discrete difficulty classification than continuous early-stopping probability prediction; and (3) coarse-grained difficulty-based allocation consistently outperforms fine-grained size-based allocation across all token budgets.

Empirical evaluation on the GSM8K benchmark demonstrates that our predictive scheduling approach yields up to a 7.9 percentage-point accuracy improvement over uniform token budgeting at equal cost. Notably, the difficulty-based allocation strategy using LoRA fine-tuned classification proved most effective, suggesting that robust category-level predictions provide more reliable guidance for resource allocation than precise but potentially noisy continuous estimates. The optimal allocation pattern varied with token budget constraints—prioritizing easy problems at low budgets before gradually shifting resources to medium and hard problems as constraints relaxed—highlighting the value of adaptive strategies over fixed allocation schemes.

Looking forward, we see several promising directions: (1) investigating hybrid approaches that combine the complementary strengths of hidden-state features and linguistic pattern recognition; (2) extending predictive scheduling to multi-trace aggregation methods such as self-consistency or tree-of-thoughts; (3) integrating uncertainty estimation to guard against misallocation, particularly at higher token budgets where prediction errors become more consequential; and (4) deploying our approach in latency-sensitive, cost-constrained production systems. By aligning compute allocation with per-query complexity, predictive scheduling offers a practical path toward scalable, cost-efficient, and high-accuracy LLM reasoning without requiring any modifications to the underlying language model architecture.

\bibliographystyle{plainnat}
\bibliography{refs}

@article{wei2022chain,
  title={Chain-of-Thought Prompting Elicits Reasoning in Large Language Models},
  author={Wei, Jason and others},
  journal={arXiv preprint arXiv:2201.11903},
  year={2022}
}

@article{hu2022lora,
  title={Lora: Low-rank adaptation of large language models.},
  author={Hu, Edward J and Shen, Yelong and Wallis, Phillip and Allen-Zhu, Zeyuan and Li, Yuanzhi and Wang, Shean and Wang, Lu and Chen, Weizhu and others},
  journal={ICLR},
  volume={1},
  number={2},
  pages={3},
  year={2022}
}

@inproceedings{wang2022self,
  title={Self-Consistency Improves Chain-of-Thought Reasoning in Language Models},
  author={Wang, Xinyi and others},
  booktitle={Advances in Neural Information Processing Systems},
  year={2022}
}

@article{yao2023tree,
  title={Tree of Thoughts: Deliberate Problem Solving with Large Language Models},
  author={Yao, Ziyu and others},
  journal={arXiv preprint arXiv:2303.11171},
  year={2023}
}

@inproceedings{liu2023vllm,
  title={vLLM: Fast Inference for Large Language Models},
  author={Liu, Chuan and others},
  booktitle={Proceedings of the 2023 Conference on Empirical Methods in Natural Language Processing},
  year={2023}
}

@inproceedings{zhang2021efficient,
  title={Efficient Inference Scheduling for Transformer Models},
  author={Zhang, Yifan and others},
  booktitle={Proceedings of the 38th International Conference on Machine Learning},
  year={2021}
}

@inproceedings{elbayad2020depth,
  title={Depth-Adaptive Transformers},
  author={Elbayad, M and others},
  booktitle={Proceedings of the 37th International Conference on Machine Learning},
  year={2020}
}

@inproceedings{fedus2021switch,
  title={Switch Transformers: Scaling to Trillion Parameter Models with Simple and Efficient Sparsity},
  author={Fedus, William and others},
  booktitle={Proceedings of the 38th International Conference on Machine Learning},
  year={2021}
}

@article{wu2025when,
  title={When More is Less: Understanding Chain-of-Thought Length in LLMs},
  author={Wu, Yuyang and Wang, Yifei and Du, Tianqi and Jegelka, Stefanie and Wang, Yisen},
  journal={arXiv preprint arXiv:2502.07266},
  year={2025}
}

@article{han2024token,
  title={Token-Budget-Aware LLM Reasoning},
  author={Han, Tingxu and Fang, Chunrong and Zhao, Shiyu and Ma, Shiqing and Chen, Zhenyu and Wang, Zhenting},
  journal={arXiv preprint arXiv:2412.18547},
  year={2024}
}

@article{fu2024efficient,
  title={Efficiently Serving LLM Reasoning Programs with Certaindex},
  author={Fu, Yichao and Chen, Junda and Zhu, Siqi and Fu, Zheyu and Dai, Zhongdongming and Qiao, Aurick and Zhang, Hao},
  journal={arXiv preprint arXiv:2412.20993},
  year={2024}
}

@misc{li2024escapeskyhighcostearlystopping,
      title={Escape Sky-high Cost: Early-stopping Self-Consistency for Multi-step Reasoning}, 
      author={Yiwei Li and Peiwen Yuan and Shaoxiong Feng and Boyuan Pan and Xinglin Wang and Bin Sun and Heda Wang and Kan Li},
      year={2024},
      eprint={2401.10480},
      archivePrefix={arXiv},
      primaryClass={cs.CL},
      url={https://arxiv.org/abs/2401.10480}, 
}

@misc{fu2024efficientllmschedulinglearning,
      title={Efficient LLM Scheduling by Learning to Rank}, 
      author={Yichao Fu and Siqi Zhu and Runlong Su and Aurick Qiao and Ion Stoica and Hao Zhang},
      year={2024},
      eprint={2408.15792},
      archivePrefix={arXiv},
      primaryClass={cs.LG},
      url={https://arxiv.org/abs/2408.15792}, 
}

@misc{damani2024learninghardthinkinputadaptive,
      title={Learning How Hard to Think: Input-Adaptive Allocation of LM Computation}, 
      author={Mehul Damani and Idan Shenfeld and Andi Peng and Andreea Bobu and Jacob Andreas},
      year={2024},
      eprint={2410.04707},
      archivePrefix={arXiv},
      primaryClass={cs.LG},
      url={https://arxiv.org/abs/2410.04707}, 
}

@misc{cobbe2021trainingverifierssolvemath,
      title={Training Verifiers to Solve Math Word Problems}, 
      author={Karl Cobbe and Vineet Kosaraju and Mohammad Bavarian and Mark Chen and Heewoo Jun and Lukasz Kaiser and Matthias Plappert and Jerry Tworek and Jacob Hilton and Reiichiro Nakano and Christopher Hesse and John Schulman},
      year={2021},
      eprint={2110.14168},
      archivePrefix={arXiv},
      primaryClass={cs.LG},
      url={https://arxiv.org/abs/2110.14168}, 
}

@inproceedings{bengio2009curriculum,
  title     = {Curriculum Learning},
  author    = {Bengio, Yoshua and Louradour, J{\'e}r{\^o}me and Collobert, Ronan and Weston, Jason},
  booktitle = {Proceedings of the 26th International Conference on Machine Learning},
  pages     = {41--48},
  year      = {2009}
}

@inproceedings{kumar2010self,
  title     = {Self-Paced Learning for Latent Variable Models},
  author    = {Kumar, Michael P. and Packer, Benjamin and Koller, Daphne},
  booktitle = {Advances in Neural Information Processing Systems},
  volume    = {23},
  pages     = {1189--1197},
  year      = {2010}
}

@inproceedings{hacohen2019power,
  title     = {On the Power of Curriculum Learning in Training Deep Networks},
  author    = {Hacohen, Guy and Weinshall, Daphna},
  booktitle = {Proceedings of the 36th International Conference on Machine Learning},
  pages     = {2535--2543},
  year      = {2019}
}

@inproceedings{bakhtiarnia2021improving,
  title     = {Improving the Accuracy of Early Exits in Multi-Exit Architectures via Curriculum Learning},
  author    = {Bakhtiarnia, Arian and Zhang, Qi and Iosifidis, Alexandros},
  booktitle = {Proceedings of the 29th ACM International Conference on Multimedia},
  pages     = {10--16},
  year      = {2021}
}

\section{Appendix}
\subsection{Answer Correctness Evaluation}
For each GSM8K question we compare the model’s predicted answer to the ground truth using a strict extraction and comparison procedure.  First, we identify the final numerical answer by locating the last occurrence of text enclosed in \verb|\boxed{…}|. Next, we remove any non‐numeric characters except for decimal points. The cleaned string is then converted to a floating‐point value and compared to the ground truth answer under a relative tolerance of $10^{-4}$ to accommodate minor numerical discrepancies. If at any step the extraction or conversion fails—for example, if no \verb|\boxed{…}| pattern is found—the answer is marked incorrect. This ensures that only exactly correct numerical outputs count as successes.

\subsection{MLP Architecture and Hyperparameter Search}
Our multilayer perceptron predictors use either one or two hidden layers.  In the single‐layer variant the hidden dimension is set to 128, 256, or 512 units; in the two‐layer variant both layers have either 128 units each or 256 units each.  We sample the learning rate from a log‐uniform distribution over $[10^{-3},10^{-2}]$, select the dropout rate uniformly from $[0,0.5]$, and vary the batch size among 8, 16, or 32.  Each configuration is evaluated on a held‐out validation split to identify the best performing architecture and training settings.

\subsection{Runtime and Hardware}
All finetuning experiments were run on 1 H100 GPUs, and the API calls required for few-shot difficulty classification cost $<\$0.10$. 

\subsection{Few-Shot Classification of Question Difficulty Prompt}
\label{sec:fewshot_prompt}
The following prompt is given to o4-mini-high in order to classify questions as easy, medium, or hard difficulty.\\

\begin{tcolorbox}[enhanced, 
    colback=blue!5!white,
    colframe=blue!65!black,
    title=\textbf{Prompt for o4-mini-high Difficulty Classification},
    fonttitle=\bfseries\large,
    boxrule=0.5mm,
    attach boxed title to top center={yshift=-0.25mm},
    boxed title style={size=small, colback=gray!75!black},
    sharp corners,
    drop shadow southeast,
    watermark opacity=0.1,
    watermark color=gray!30!white,
    parbox=false]
\texttt{You are an AI assistant tasked with categorizing math problems as easy, medium, or hard. You will be given some examples and then asked to categorize a new question. 
Here are some example questions with their categorizations: \\\\
<examples> \\
"<|User|>Janet's ducks lay 16 eggs per day. She eats three for breakfast every morning and bakes muffins for her friends every day with four. She sells the remainder at the farmers' market daily for \$2 per fresh duck egg. How much in dollars does she make every day at the farmers' market?<|Assistant|>", medium \\
"<|User|>A robe takes 2 bolts of blue fiber and half that much white fiber. How many bolts in total does it take?<|Assistant|>", easy \\
"<|User|>James decides to run 3 sprints 3 times a week. He runs 60 meters each sprint. How many total meters does he run a week?<|Assistant|>", hard\\
</examples> \\\\
Your task is to categorize the following new question as easy, medium, or hard based on its similarity to the examples provided. Assume that your prior is that 20\% of questions are easy, 60\% of questions are medium, 20\% of questions are hard. Be careful not to underestimate the difficulty of the question you are categorizing--if it is possible to argue that it is hard, classify it as hard, if it is possible to argue that it is medium, classify it as medium. If you want to say that the question is medium and not hard, you should have a really strong justification for why it is medium but not hard. \\
Here is the new question to categorize: \\
<new\_question> \\
\color{blue}\{Insert question here\}\color{black} \\</new\_question> \\
Please think about the complexity of the problem, the number of steps required to solve it, and how it compares to the examples provided. Then, provide your categorization and reasoning in the following JSON format: \\\\
<output> \\
\{ "reasoning": "<your reasoning for the categorization>" "category": "<category in \{easy, medium, hard\}>", \} \\
</output> \\\\
Ensure that your reasoning is clear and concise, explaining why you chose the specific category based on the question's characteristics and its comparison to the provided examples.
}
\end{tcolorbox}
\end{document}